\documentclass[runningheads]{llncs}
\usepackage{graphicx}
\usepackage{amsmath,amssymb} 
\usepackage{color}
\usepackage[width=122mm,left=12mm,paperwidth=146mm,height=193mm,top=12mm,paperheight=217mm]{geometry}
\usepackage{epsfig}
\usepackage{subcaption}
\usepackage{indentfirst}
\usepackage{multirow}
\usepackage{booktabs}
\usepackage{threeparttable}
\usepackage{boldline}
\usepackage{tabularx}
\usepackage{makecell}
\usepackage{amsmath}
\usepackage{breqn}
\usepackage{booktabs,amsfonts,dcolumn}

\usepackage[breaklinks=true,bookmarks=false]{hyperref}

\DeclareMathOperator*{\vect}{vec}

\usepackage{floatrow}
\newfloatcommand{capbtabbox}{table}[][\FBwidth]

\begin{document}
\pagestyle{headings}

\newcommand{\A}{\mathbf{A}}
\newcommand{\B}{\mathbf{B}}
\newcommand{\C}{\mathbf{C}}
\newcommand{\D}{\mathbf{D}}
\newcommand{\E}{\mathbf{E}}
\newcommand{\F}{\mathbf{F}}
\newcommand{\G}{\mathbf{G}}
\newcommand{\bH}{\mathbf{H}}
\newcommand{\I}{\mathbf{I}}
\newcommand{\J}{\mathbf{J}}
\newcommand{\K}{\mathbf{K}}
\newcommand{\bL}{\mathbf{L}}
\newcommand{\M}{\mathbf{M}}
\newcommand{\bO}{\mathbf{O}}
\newcommand{\bP}{\mathbf{P}}
\newcommand{\Q}{\mathbf{Q}}
\newcommand{\R}{\mathbf{R}}
\newcommand{\bS}{\mathbf{S}}
\newcommand{\T}{\mathbf{T}}
\newcommand{\U}{\mathbf{U}}
\newcommand{\V}{\mathbf{V}}
\newcommand{\W}{\mathbf{W}}
\newcommand{\X}{\mathbf{X}}
\newcommand{\Y}{\mathbf{Y}}
\newcommand{\Z}{\mathbf{Z}}

\newcommand{\ba}{\mathbf{a}}
\newcommand{\bb}{\mathbf{b}}
\newcommand{\bc}{\mathbf{c}}
\newcommand{\bd}{\mathbf{d}}
\newcommand{\be}{\mathbf{e}}
\newcommand{\f}{\mathbf{f}}
\newcommand{\g}{\mathbf{g}}
\newcommand{\h}{\mathbf{h}}
\newcommand{\bk}{\mathbf{k}}
\newcommand{\m}{\mathbf{m}}
\newcommand{\n}{\mathbf{n}}
\newcommand{\p}{\mathbf{p}}
\newcommand{\q}{\mathbf{q}}
\newcommand{\br}{\mathbf{r}}
\newcommand{\s}{\mathbf{s}}
\newcommand{\bt}{\mathbf{t}}
\newcommand{\bu}{\mathbf{u}}
\newcommand{\bv}{\mathbf{v}}
\newcommand{\w}{\mathbf{w}}
\newcommand{\x}{\mathbf{x}}
\newcommand{\y}{\mathbf{y}}
\newcommand{\z}{\mathbf{z}}

\newcommand{\cA}{\mathcal{A}}
\newcommand{\cB}{\mathcal{B}}
\newcommand{\cC}{\mathcal{C}}
\newcommand{\cD}{\mathcal{D}}
\newcommand{\cF}{\mathcal{F}}
\newcommand{\cG}{\mathcal{G}}
\newcommand{\cH}{\mathcal{H}}
\newcommand{\cI}{\mathcal{I}}
\newcommand{\cL}{\mathcal{L}}
\newcommand{\cM}{\mathcal{M}}
\newcommand{\cN}{\mathcal{N}}
\newcommand{\cP}{\mathcal{P}}
\newcommand{\cQ}{\mathcal{Q}}
\newcommand{\cR}{\mathcal{R}}
\newcommand{\cS}{\mathcal{S}}
\newcommand{\cT}{\mathcal{T}}
\newcommand{\cU}{\mathcal{U}}
\newcommand{\cV}{\mathcal{V}}
\newcommand{\cX}{\mathcal{X}}
\newcommand{\cY}{\mathcal{Y}}
\newcommand{\cZ}{\mathcal{Z}}

\newcommand{\bp}{\bar{\p}}
\newcommand{\BP}{\bar{\bP}}

\def\bphi{\mbox{\boldmath $\phi$}}
\def\bdelta{\mbox{\boldmath $\delta$}}
\def\blambda{\mbox{\boldmath $\lambda$}}
\def\bLambda{\mbox{\boldmath $\Lambda$}}
\def\bPi{\mbox{\boldmath $\Pi$}}
\newcommand{\bTheta}{\mathbf{\Theta}}
\newcommand{\bSigma}{\mathbf{\Sigma}}
\newcommand{\one}{\mathbf{1}}
\newcommand{\zero}{\mathbf{0}}
\newcommand{\epi}{\mathbf{epi}}
\newcommand{\dom}{\mathbf{dom}}
\newcommand{\real}{\mathbb{R}}
\newcommand{\bina}{\{0, 1\}}
\newcommand{\integer}{\mathbb{Z}}
\newcommand{\acc}{\text{acc}}
\newcommand{\err}{\text{err}}
\newcommand{\dist}{\text{dist}}
\newcommand{\convA}{\stackbin[\A]{}{*}}
\newcommand{\convAj}{\stackbin{\A_j}{*}}
\newcommand{\Searrow}{\rotatebox[origin=c]{-45}{$\Rightarrow$}}
\newcommand{\udots}{\mathinner{\mskip1mu\raise1pt\vbox{\kern7pt\hbox{.}}\mskip2mu\raise4pt\hbox{.}\mskip2mu\raise7pt\hbox{.}\mskip1mu}}

\def\astm{\ \dot{*}_m  }
\def\astmm{\ \ddot{*}_m  }
\def\astl{\ \dot{*}_l \ }
\def\astll{\ \ddot{*}_l \ }
\def\astx{ \dot{*}_{n_x}  }
\def\astxx{ \ddot{*}_{n_x}  }
\def\asts{ \ \dot{*}_{n_s} }
\def\astss{ \ \ddot{*}_{n_s} }
\def\astsss{ \ \dddot{*}_{n_s} }
\def\astz{ \ \dot{*}_{n_z}  }
\def\astzz{ \ \ddot{*}_{n_z}  }
\renewcommand\arraystretch{0.8}

\newcommand{\Hl}{\overleftarrow{\bH}}
\newcommand{\Hr}{\overrightarrow{\bH}}
\newcommand{\dk}{\dot{k}}
\newcommand{\dd}{\dot{d}}
\newcommand{\hQ}{\hat{\Q}}


\newcommand\etal{\emph{et al.}}
\newcommand\eg{\emph{e.g.}}
\newcommand\ie{\emph{i.e.}}
\newcommand\wrt{\emph{w.r.t.}}
\newcommand\aka{\emph{a.k.a.}}
\newenvironment{aligns}{\par\nobreak\small\noindent\align}{\endalign}
\newcommand{\squeezeup}{\vspace{-1.5mm}}

\setlength{\abovecaptionskip}{0pt}
\setlength{\belowcaptionskip}{0pt}

\mainmatter
\def\ECCV18SubNumber{2343}  

\title{Multi-Attention Multi-Class Constraint for Fine-grained Image Recognition} 

\titlerunning{Multi-Attention Multi-Class Constraint for Fine-grained Image Recognition}

\authorrunning{Ming Sun, Yuchen Yuan, Feng Zhou, and Errui Ding}

\author{Ming Sun, Yuchen Yuan, Feng Zhou*, and Errui Ding}


\institute{Department of Computer Vision Technology (VIS), 
	Baidu Inc.\\
	\email{ \{sunming05, yuanyuchen02, zhoufeng09, dingerrui\}@baidu.com}
}

\maketitle

\vspace{-0.2in}
\begin{abstract}
Attention-based learning for fine-grained image recognition remains a challenging task, where most of the existing methods treat each object part in isolation, while neglecting the correlations among them. In addition, the multi-stage or multi-scale mechanisms involved make the existing methods less efficient and hard to be trained end-to-end.
In this paper, we propose a novel attention-based convolutional neural network (CNN) which regulates multiple object parts among different input images.
Our method first learns multiple attention region features of each input image through the one-squeeze multi-excitation (OSME) module, and then apply the multi-attention multi-class constraint (MAMC) in a metric learning framework.
For each anchor feature, the MAMC functions by pulling same-attention same-class features closer, while pushing different-attention or different-class features away.
Our method can be easily trained end-to-end, and is highly efficient which requires only one training stage.
Moreover, we introduce Dogs-in-the-Wild, a comprehensive dog species dataset that surpasses similar existing datasets by category coverage, data volume and annotation quality.
This dataset will be released upon acceptance to facilitate the research of fine-grained image recognition.
Extensive experiments are conducted to show the substantial improvements of our method on four benchmark datasets.
\end{abstract}

\section{Introduction}
\label{introduction}

\begin{figure}
\centering
\includegraphics[width=0.7\textwidth]{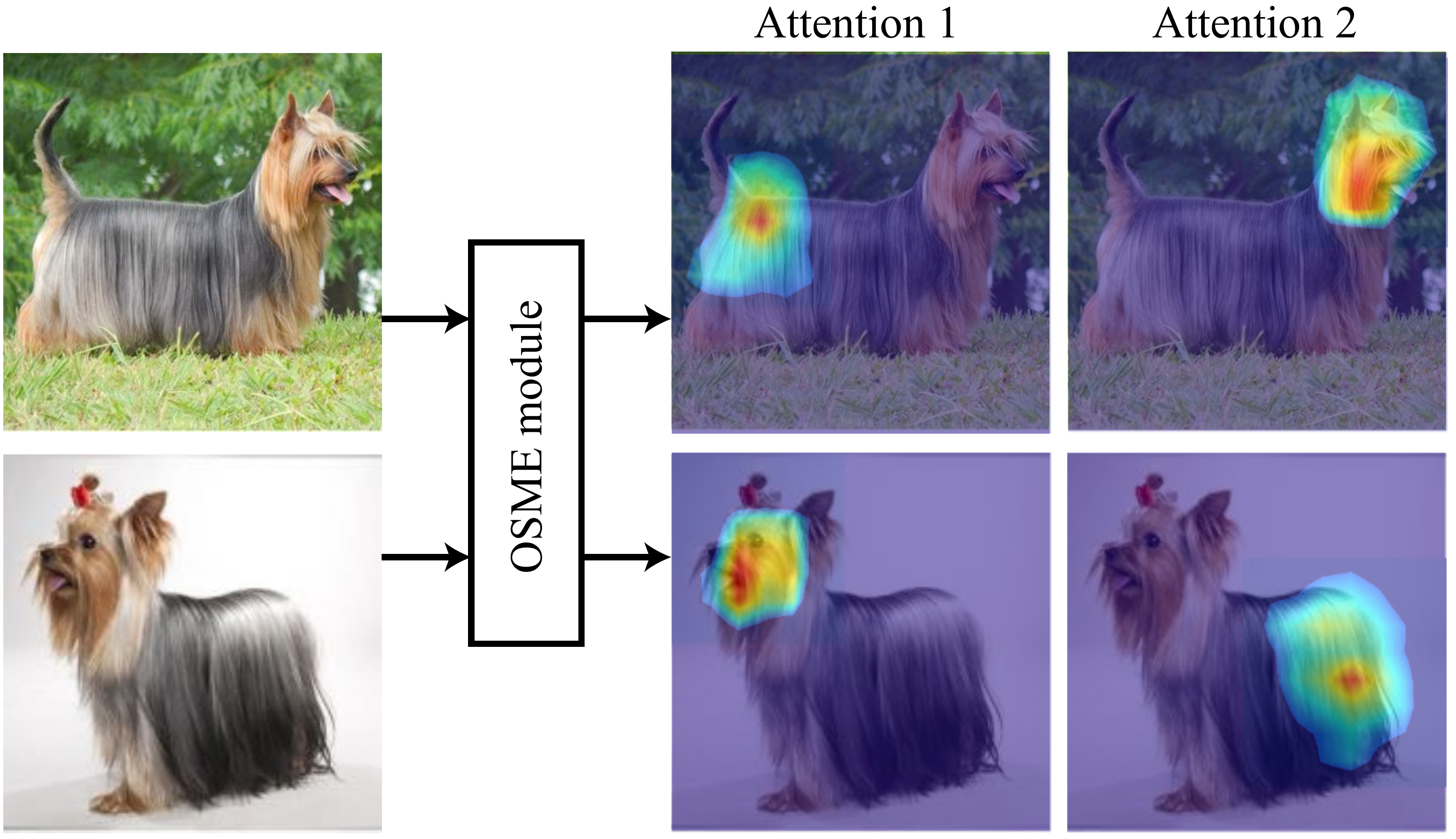}
\caption{Two distinct dog species from the proposed Dogs-in-the-Wild dataset. Our method is capable of capturing the subtle differences on the head and tail without manual part annotations.}
\label{fig:teaser}
\vspace{-.1in}
\end{figure}

In the past few years, the performances of generic image recognition on large-scale datasets (\eg, ImageNet \cite{deng2009imagenet}, Places~\cite{ZhouLXTO14}) have undergone unprecedented improvements, thanks to the breakthroughs in the design and training of deep neural networks (DNNs).
Such fast-pacing progresses in research have also drawn attention of the related industries to build software like Google Lens on smartphones to recognize everything snapshotted by the user.
Yet, recognizing the fine-grained category of daily objects such as car models, animal species or food dishes is still a challenging task for existing methods.
The reason is that the global geometry and appearances of fine-grained classes can be very similar, and how to identify their subtle differences on the key parts is of vital importance.
For instance, to differentiate the two dog species in Figure~\ref{fig:teaser}, it is important to consider their discriminative features on the ear, tail and body length, which is extremely difficult to notice even for human without domain expertise.

Thus the majority of efforts in the fine-grained community focus on how to effectively integrate part localization into the classification pipeline.
In the pre-DNN era, various parametric~\cite{farrell2011birdlets,LinMHD14,parkhi2011truth} and non-parametric~\cite{LiuKJB12} part models have been employed to extract discriminative part-specific features.
Recently, with the popularity of DNNs, the tasks of object part localization and feature representation can be both learned in a more effective way \cite{lin2015deep,krause2015fine,branson2014bird,zhang2014part,zhang2014panda}.
The major drawback of these strongly-supervised methods, however, is that they heavily rely on manual object part annotations, which is too expensive to be prevalently applied in practice.
Therefore, weakly-supervised frameworks have received increasing attention in recent researches.
For instance, the attention mechanism can be implemented as sequential decision processes~\cite{MnihHGK14} or multi-stream part selections~\cite{fu2017look} without the need of part annotations.
Despite the great progresses, these methods still suffer several limitations.
First, their additional steps, such as the part localization and feature extraction of the attended regions, can incur expensive computational cost.
Second, their training procedures are sophisticated, requiring multiple alternations or cascaded stages due to the complex architecture designs.
More importantly, most works tend to detect the object parts in isolation, while neglect their inherent correlations.
As a consequence, the learned attention modules are likely to focus on the same region and lack the capability to localize multiple parts with discriminative features that can differentiate between similar fine-grained classes.

From extensive experimental studies, we observe that an effective visual attention mechanism for fine-grained classification should follow three criteria:
1) The detected parts should be well spread over the object body to extract non-correlated features;
2) Each part feature alone should be discriminative for separating objects of different classes;
3) The part extractors should be lightweight in order to be scaled up for practical applications.
To meet these demands, this paper presents a novel framework that contains two major improvements.
First, we propose one-squeeze multi-excitation module (OSME) to localize different parts inspired by the latest ImageNet winner SENet~\cite{hu2017squeeze}.
OSME is a fully differentiable unit and is capable of directly extracting part features with budgeted computational cost, unlike existing methods that explicitly cropping the object part first and then feedforward again for the feature.
Second, inspired by metric learning loss, we propose the multi-attention multi-class constraint (MAMC) to coherently enforce the correlations among different parts in the training of fine-grained object classifiers. MAMC encourages same-attention same-class features to be closer than different-attention or different-class ones.
In addition, we have collected a new dataset of dog species called Dogs-in-the-Wild, which exhibits higher category coverage, data volume and annotation quality than similar public datasets.
Experimental results show that our method achieves substantial improvements on four benchmark datasets.
Moreover, our method can be easily trained end-to-end, and unlike most existing methods that require multiple feedforward processes for feature extraction \cite{wang2017residual,zhao2016diversified} or multiple alternative training stages \cite{fu2017look,rosenfeld2016visual}, only one stage and one feedforward are required for each training step of our network, which offers significantly improved efficiency.


\vspace{-.1in}
\section{Related Work}
\label{related_work}

\subsection{Fine-Grained Image Recognition}
In the task of fine-grained image recognition, since the inter-class differences are subtle, more specialized techniques, including discriminative feature learning and object parts localization, need to be applied.
A straightforward way is supervised learning with manual object part annotations, which has shown promising results in classifying birds~\cite{branson2014bird,farrell2011birdlets,zhang2014part,zhang2014panda}, dogs~ \cite{khosla2011novel,parkhi2011truth,LiuKJB12,zhang2014part}, and cars~\cite{krause2014learning,LinMHD14,krause20133d}.
However, it is usually laborious and expensive to obtain object part annotations, which severely restricts the effectiveness of such methods.

Consequently, more recently proposed methods tend to localize object parts with weakly-supervised mechanisms,
such as the combination of pose alignment and co-segmentation \cite{krause2015fine}, dynamic spatial transformation of the input image for better alignment \cite{jaderberg2015spatial}, and parallel CNNs for bilinear feature extraction \cite{lin2015bilinear}.
Compared with previous works, our method also takes a weakly-supervised mechanism, but can directly extract the part features without cropping them out, and is highly efficient to be scaled up with multiple parts.

In recent years, more advanced methods emerge with improved results.
For instance, the bipartite-graph labeling \cite{zhou2016fine} leverages the label hierarchy on the fine-grained classes, which is less expensive to obtain.
The work in \cite{zhang2016picking} exploit unified CNN framework with spatially weighted representation by the Fisher vector \cite{perronnin2015fisher}.
\cite{branson2014ignorant} and \cite{wilber2015learning} incorporate human knowledge and various types of computer vision algorithms into a human-in-the-loop framework for the complementary strengths of both ends.
And in \cite{simon2017generalized}, the average and bilinear pooling are combined to learn the pooling strategy during training.
These techniques can also be potentially combined with our method for further works.

\vspace{-.1in}
\subsection{Visual Attention}
The aforementioned part-based methods have shown strong performances in fine-grained image recognition.
Nevertheless, one of their major drawbacks is that they need meaningful definitions of the object parts, which are hard to obtain for non-structured objects such as flowers~\cite{NilsbackZ08} and food dishes~\cite{BossardGG14}.
Therefore, the methods enabling CNN to attend loosely defined regions for general objects have emerged as a promising direction.
For instance, the soft proposal network \cite{zhu2017soft} combines random walk and CNN for object proposals.
The works in \cite{zhao2016diversified} and \cite{liu2016fully} introduce long short-term memory~\cite{hochreiter1997long} and reinforcement learning to attention-based classification, respectively.
And the class activation mapping \cite{zhou2016learning} generates the heatmap of the input image, which provides a better way for attention visualization.
On the other hand, the idea of multi-scale feature fusion or recurrent learning has become increasingly popular in recent works.
For instance, the work in \cite{rosenfeld2016visual} extends \cite{zhou2016learning} and establishes a cascaded multi-stage framework, which refines the attention region by iteration.
The residual attention network \cite{wang2017residual} obtains the attention mask of input image by up-sampling and down-sampling, and a series of such attention modules are stacked for feature map refinement.
And the recurrent attention CNN \cite{fu2017look} alternates between the optimization of softmax and pairwise ranking losses, which jointly contribute to the final feature fusion.
Even an acceleration method \cite{li2017dynamic} with reinforcement learning is proposed particularly for the recurrent attention models above.

In parallel to these efforts, our method not only automatically localizes the attention regions, but also directly captures the corresponding features without explicitly cropping the ROI and feedforwarding again for the feature, which makes our method highly efficient.

\vspace{-.1in}
\subsection{Metric Learning}

Apart from the techniques above, deep metric learning aims at the learning of appropriate similarity measurements between sample pairs, which provides another promising direction to fine-grained image recognition.
Classical metric learning may be considered as learning of the Mahalanobis distance between pairs of points \cite{kulis2013metric}.
The pioneer work of Siamese network~\cite{bromley1994signature} formulates the deep metric learning with a contrastive loss that minimizes distance between positive pairs while keeps negative pairs apart.
Despite its great success on face verification~\cite{SchroffKP15}, contrastive embedding requires that training data contains real-valued precise pair-wise similarities or distances.
The triplet loss \cite{salakhutdinov2007learning} addresses this issue by optimizing the relative distance of the positive pair and one negative pair from three samples.
It has been proven that triplet loss is extremely effective for fine-grained product search~\cite{WangSLRWPCW14}.
Later, triplet loss is improved to automatically search for discriminative patches~\cite{wang2016mining}.
Nevertheless, compared with softmax loss, triplet loss is difficult to train due to its slow convergence.
To alleviate this issue, the N-pair loss~\cite{sohn2016improved} is introduced to consider multiple negative samples in training, and exhibits higher efficiency and performance.
More recently, the angular loss~\cite{wang2017deep} enhances N-pair loss by integrating high-order constraint that captures additional local structure of triplet triangles.

Our method differs previous metric learning works in two aspects:
First, we take object parts instead of the whole images as instances in the feature learning process;
Second, our formulation simultaneously considers the part and class labels of each instance.

\begin{figure*}[t]
  \centering
  \includegraphics[width=\textwidth]{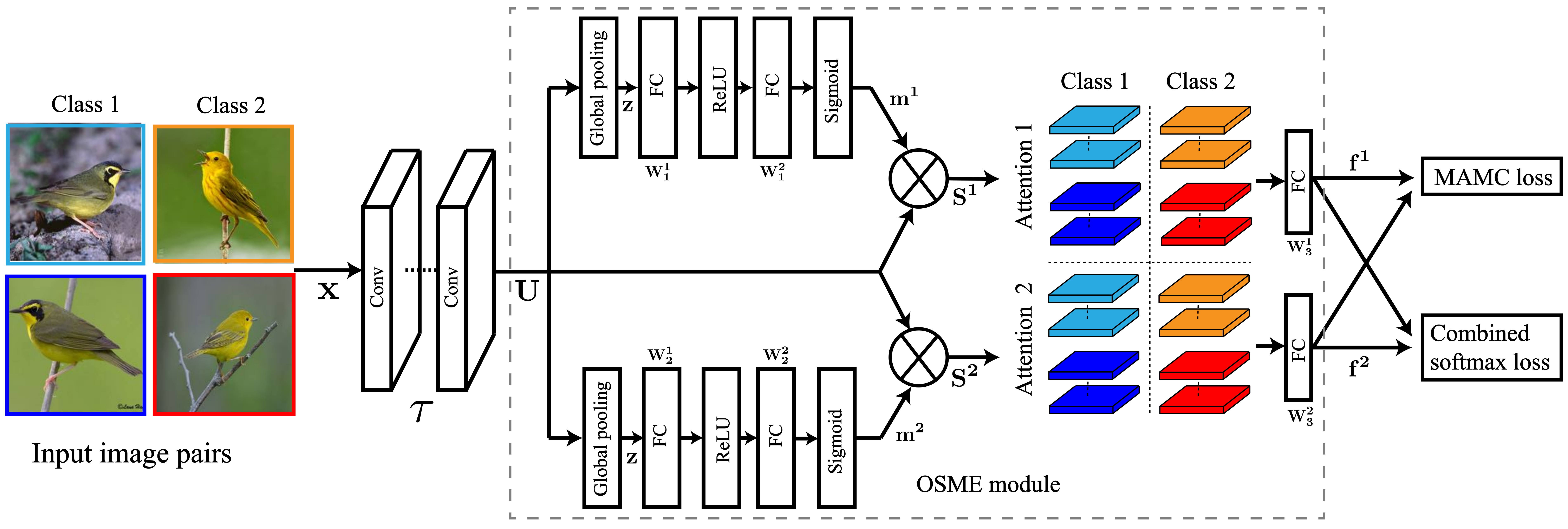}
  \caption{Overview of our network architecture.
    Here we visualize the case of learning two attention branches given a training batch with four images of two classes.
    The MAMC and softmax losses would be replaced by a softmax layer in testing.
    Unlike hard-attention methods like~\cite{fu2017look}, we do not explicitly crop the parts out.
    Instead, the feature maps ($\bS^1$ and $\bS^2$) generated by the two branches provide soft response for attention regions such as the birds' head or torso, respectively.}
  \label{fig:flowchart_all}
\end{figure*}

\vspace{-.1in}
\section{Proposed Method}
\label{approach}

In this section, we present our proposed method which can efficiently and accurately attend discriminative regions despite being trained only on image-level labels.
As shown in Figure~\ref{fig:flowchart_all}, the framework of our method is composed by two parts:
1) A differentiable one-squeeze multi-excitation (OSME) module that extracts features from multiple attention regions with a slight increase in computational burden.
2) A multi-attention multi-class (MAMC) constraint that enforces the correlation of the attention features in favor of the fine-grained classification task.
In contrast to many prior works, the entire network of our method can be effectively trained end-to-end in one stage.


\vspace{-.1in}
\subsection{One-Squeeze Multi-Excitation Attention Module}
\label{osme}

There have been a number of visual attention models exploring weakly supervised part localization, and the previous works can be roughly categorized in two groups.
The first type of attention is also known as part detection, \ie, each attention is equivalent to a bounding box covering a certain area.
Well-known examples include the early work of recurrent visual attention~\cite{MnihHGK14}, the spatial transformer networks~\cite{jaderberg2015spatial}, and the recent method of recurrent attention CNN~\cite{fu2017look}.
This hard-attention setup can benefit a lot from the object detection community in the formulation and training.
However, its architectural design is often cumbersome as the part detection and feature extraction are separated in different modules.
For instance, the authors of \cite{jaderberg2015spatial} apply three GoogLeNets~\cite{SzegedyLJSRAEVR15} for detecting and representing two parts of birds.
As the base network goes deeper, the memory and computational cost would become too high to afford for more than three object parts even using the latest GPUs.
The second type of attention can be considered as imposing a soft mask on the feature map, which origins from activation visualization~\cite{ZeilerF14,ZhouKLOT14}.
Later, people find it can be extended for localizing parts~\cite{zhou2016learning,rosenfeld2016visual} and improving the overall recognition performance~\cite{wang2017residual,hu2017squeeze}.
Our approach also falls into this category.
We adopt the idea of SENet~\cite{hu2017squeeze}, the latest ImageNet winner, to capture and describe multiple discriminative regions in the input image.
Compared to other soft-attention works~\cite{zhou2016learning,wang2017residual}, we build on SENet because of its superiority in performance and scalability in practice.

As shown in Figure~\ref{fig:flowchart_all}, our framework is a feedforward neural network where each image is first processed by a base network, \eg, ResNet-50~\cite{he2016deep}.
Let $\x \in \real^{W' \times H' \times C'}$ denote the input fed into the last residual block $\tau$.
The goal of SENet is to re-calibrate the output feature map,
\begin{aligns}
  \U = \tau(\x) = [ \bu_1, \cdots, \bu_C ] \in \mathbb{R}^{W \times H \times C},
\end{aligns}
through a pair of squeeze-and-excitation operations.
In order to generate $P$ attention-specific feature maps, we extend the idea of SENet by performing one-squeeze but multi-excitation operations.

In the first one-squeeze step, we aggregate the feature maps $\U$ across spatial dimensions $W \times H$ to produce a channel-wise descriptor $\z = [z_1, \cdots, z_C] \in \real^{C}$.
The global average pooling is adopted as a simple but effective way to describe each channel statistic:
\begin{aligns}
  z_c = \frac{1}{W H} \sum_{w=1}^W \sum_{h=1}^H \bu_c(w,h).
\end{aligns}
%
%

In the second multi-excitation step, a gating mechanism is independently employed on $\z$ for each attention $p = 1, \cdots, P$:
\begin{aligns}
\label{eq:w1w2}
  \m^p= \sigma \Big( \W_{2}^p \delta(\W_{1}^p \z) \Big) = [m^p_1, \cdots, m^p_C] \in \real^C,
\end{aligns}
where $\sigma$ and $\delta$ refer to the Sigmod and ReLU functions respectively.
We adopt the same design of SENet by forming a pair of dimensionality reduction and increasing layers
parameterized with $\W_{1}^p \in \mathbb{R}^{\frac{C}{r} \times {C}}$ and $\W_{2}^p \in \mathbb{R}^{{C} \times\frac{C}{r}}$.
Because of the property of the Sigmod function, each $\m^p$ encodes a non-mutually-exclusive relationship among channels.
We therefore use it to re-weight the channels of the original feature map $\U$,
\begin{aligns}
  \bS^p = [m^p_1 \bu_1, \cdots, m^p_{C} \bu_C] \in \real^{W \times H \times C}.
  \label{eq:Sp}
\end{aligns}

To extract attention-specific features, we feed each attention map $\bS^p$ to a fully connected layer $\W_3^p \in \real^{D \times WHC}$:
\begin{aligns} \label{eq:f}
  \f^p = \W_{3}^p \vect(\bS^p) \in \real^{D},
\end{aligns}
where the operator $\vect(\cdot)$ flattens a matrix into a vector.

In a nutshell, the proposed OSME module seeks to extract $P$ feature vectors $\{\f^p\}_{p=1}^P$ for each image $\x$ by adding a few layers on top of the last residual block.
Its simplicity enables the use of relatively deep base networks and an efficient one-stage training pipeline.

It is worth to clarify that the SENet is originally not designed for learning visual attentions.
By adopting the key idea of SENet, our proposed OSME module implements a lightweight yet effective attention mechanism that enables an end-to-end one-stage training on large-scale fine-grained datasets.

%

%

\vspace{-.1in}
\subsection{Multi-Attention Multi-Class Constraint}
\label{mamc}

Apart from the attention mechanism introduced in Section \ref{osme}, the other crucial problem is how to guide the extracted attention features to the correct class label.
A straightforward way is to directly evaluate the softmax loss on the concatenated attention features~\cite{jaderberg2015spatial}.
However, the softmax loss is unable to regulate the correlations between attention features.
As an alternative, another line of research~\cite{MnihHGK14,liu2016fully,fu2017look} tends to mimic human perception with a recurrent search mechanism.
These approaches iteratively generate the attention region from coarse to fine by taking previous predictions as references.
The limitation of them, however, is that the current prediction is highly dependent on the previous one, thereby the initial error could be amplified by iteration.
In addition, they require advanced techniques such as reinforcement learning or careful initialization in a multi-stage training.
In contrast, we take a more practical approach by directly enforcing the correlations between parts in training.
There has been some prior works like \cite{wang2016mining} that introduce geometrical constraints on local patches.
Our method, on the other hand, explores much richer correlations of object parts by the proposed multi-attention multi-class constraint (MAMC).


Suppose that we are given a set of training images $\{(\x, y), \cdots\}$ of $K$ fine-grained classes, where $y = 1, \cdots, K$ denotes the label associated with the image $\x$.
To model both the within-image and inter-class attention relations, we construct each training batch, $\cB = \{(\x_{i}, \x_{i}^+, y_i)\}_{i=1}^N$, by sampling $N$ pairs of images\footnote{$N$ stands for the number of sample pairs as well as the number of classes in a mini-batch. Limited by GPU memory, $N$ is usually much smaller than $K$, the total number of classes in the entire training set.} similar to~\cite{sohn2016improved}.
For each pair $(\x_i, \x_i^+)$ of class $y_i$,
the OSME module extracts $P$ attention features $\{\f_i^p, \f_i^{p+}\}_{p=1}^P$ from multiple branches according to Eq.~\ref{eq:f}.

\begin{figure}[t]
  \centering
    \vspace{-.1in}
  \includegraphics[width=0.8\textwidth]{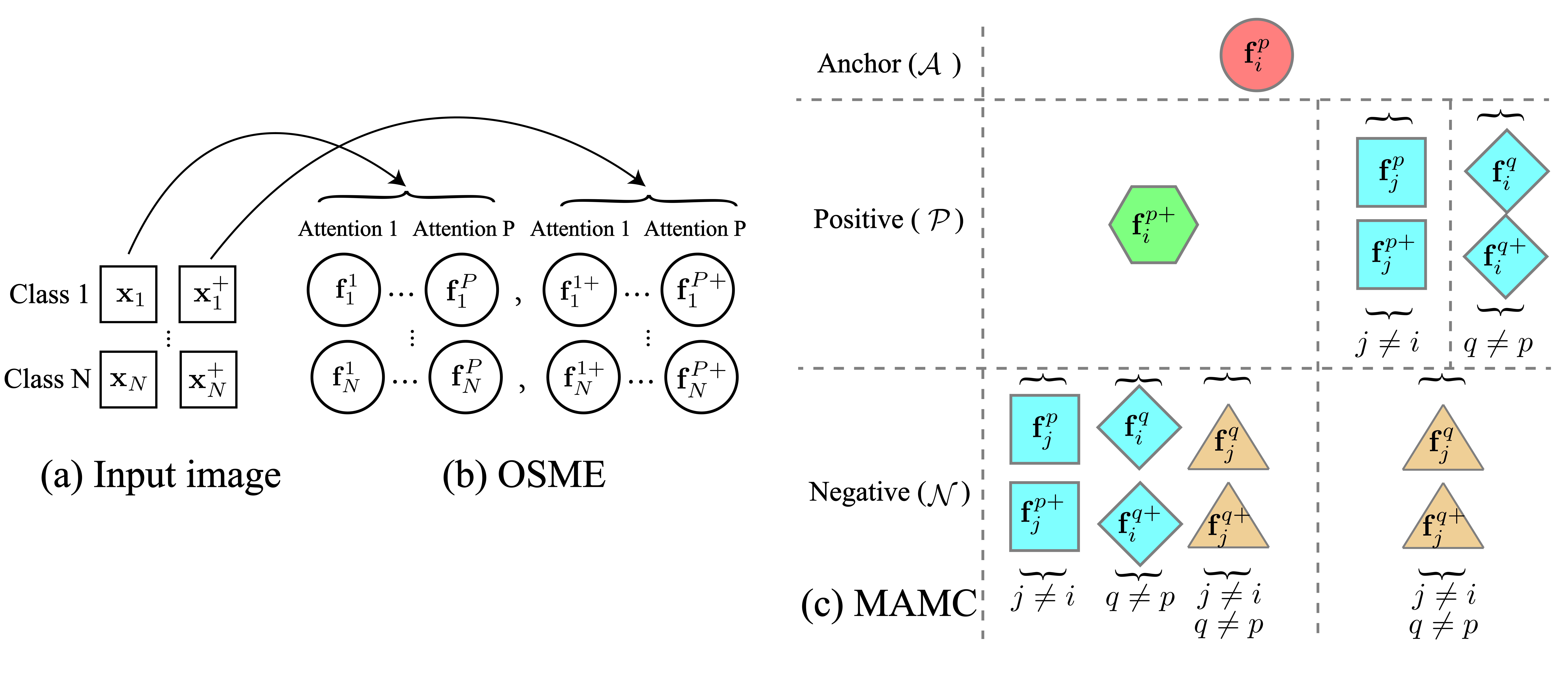}
  \vspace{-.1in}
  \caption{Data hierarchy in training. (a) Each batch is composed by $2N$ input images in N-pair style. (b) OSME extracts $P$ features for each image according to Eq.~\ref{eq:f}. (c) The group of features for three MAMC constraints by picking one feature $\f_i^p$ as the anchor.}
  \label{fig:flowchart_constraint_a}
\end{figure}

Given $2N$ samples in each batch (Figure~\ref{fig:flowchart_constraint_a}a), our intuition comes from the natural clustering of the $2NP$ features (Figure~\ref{fig:flowchart_constraint_a}b) extracted by the OSME modules.
By picking $\f_i^p$, which corresponds to the $i^{th}$ class and $p^{th}$ attention region as the anchor, we divide the rest features into four groups:
\begin{itemize}
\item same-attention same-class features, $\cS_{sasc}(\f_i^p) = \{\f_i^{p+} \}$;
\item same-attention different-class features, $\cS_{sadc}(\f_i^p) = \{ \f_j^p, \f_j^{p+} \}_{j \neq i}$;
\item different-attention same-class features, $\cS_{dasc}(\f_i^p) = \{ \f_i^{q}, \f_i^{q+} \}_{q \neq p}$;
\item different-attention different-class features $\cS_{dadc}(\f_i^p) = \{ \f_j^q, \f_j^{q+} \}_{j \neq i, q \neq p}$.
\end{itemize}

Our goal is to excavate the rich correlations among the four groups in a metric learning framework.
As summarized in Figure~\ref{fig:flowchart_constraint_a}c, we compose three types of triplets according to the choice of the positive set for the anchor $\f_i^p$.
To keep notation concise, we omit $\f_i^p$ in the following equations.

\textbf{Same-attention same-class positives.} The most similar feature to the anchor $\f_i^p$ is $\f_i^{p+}$, while all the other features should have larger distance to the anchor.
The positive and negative sets are then defined as:
\begin{aligns} \label{eq:sasc}
 \cP_{sasc} = \cS_{sasc}, \ \cN_{sasc} = \cS_{sadc} \cup \cS_{dasc} \cup \cS_{dadc}.
\end{aligns}

\textbf{Same-attention different-class positives.} For the features from different classes but extracted from the same attention region, they should be more similar to the anchor than the ones also from different attentions:
\begin{aligns} \label{eq:sadc}
 \cP_{sadc} = \cS_{sadc}, \ \cN_{sadc} = \cS_{dadc}.
\end{aligns}

\textbf{Different-attention same-class positives.} Similarly, for the features from same class but extracted from different attention regions, we have:
\begin{aligns} \label{eq:dasc}
 \cP_{dasc} = \cS_{dasc}, \ \cN_{dasc} = \cS_{dadc}.
\end{aligns}

For any positive set $\cP \in \{\cP_{sasc}, \cP_{sadc}, \cP_{dasc} \}$ and negative set $\cN \in \{\cN_{sasc}, $
$\cN_{sadc}, \cN_{dasc} \}$ combinations, we expect the anchor to be closer to the positive than to any negative by a distance margin $m > 0$, \ie,
\begin{aligns} \label{eq:triplet}
\| \f_i^p - \f^+ \|^2 + m \leq \| \f_i^p - \f^- \|^2, \ \forall \f^+ \in \cP, \f^- \in \cN.
\end{aligns}

\begin{figure}[t]
  \centering
    \vspace{-.1in}
  \includegraphics[width=0.7\textwidth]{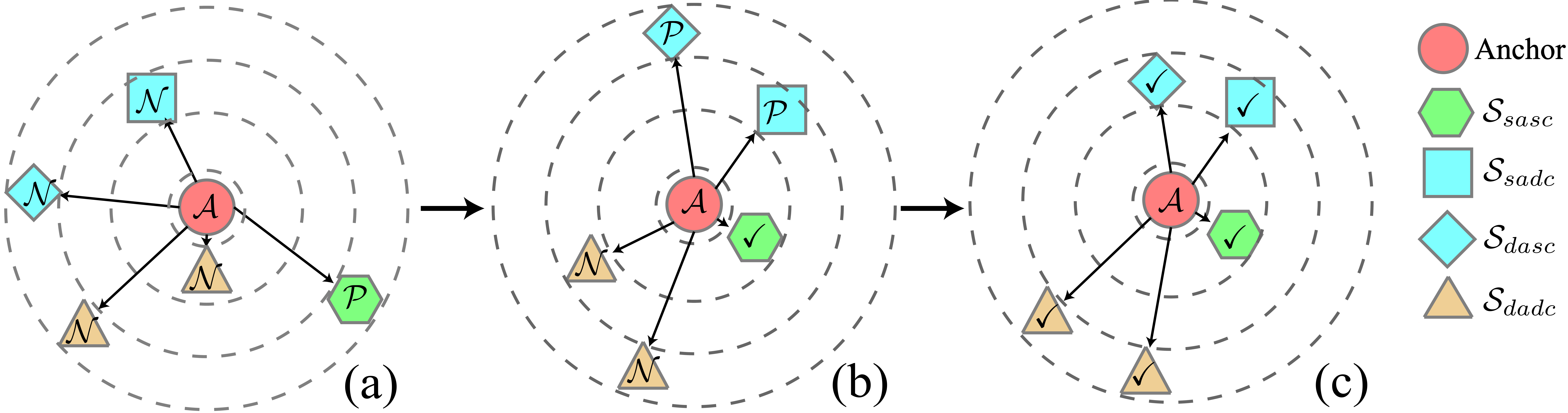}
  \caption{Feature embedding of a synthetic batch. (a) Initial embedding before learning. (b) The result embedding by applying Eq.~\ref{eq:sasc}. (c) The final embedding by enforcing Eq.~\ref{eq:sadc} and Eq.~\ref{eq:dasc}. See text for more details.}
  \label{fig:flowchart_constraint_b}
\end{figure}

To better understand the three constraints, let's consider the synthetic example of six feature points shown in Figure~\ref{fig:flowchart_constraint_b}.
In the initial state (Figure~\ref{fig:flowchart_constraint_b}a), the $\cS_{sasc}$ feature point (green hexagon) stays further away from the anchor $\f_i^p$ at the center than the others.
After applying the first constraint (Eq.~\ref{eq:sasc}), the underlying feature space is transformed to Figure~\ref{fig:flowchart_constraint_b}b, where the $\cS_{sasc}$ positive point (green \checkmark) has been pulled towards the anchor.
However, the four negative features (cyan rectangles and triangles) are still in disordered positions.
In fact, $\cS_{sadc}$ and $\cS_{dasc}$ should be considered as the positives compared to $\cS_{dadc}$ given the anchor.
By further enforcing the second (Eq.~\ref{eq:sadc}) and third (Eq.~\ref{eq:dasc}) constraints, a better embedding can be achieved in Figure~\ref{fig:flowchart_constraint_b}c, where $\cS_{sadc}$ and $\cS_{dasc}$ are regularized to be closer to the anchor than the ones of $\cS_{dadc}$.

\vspace{-.1in}
\subsection{Training Loss}

To enforce the triplet constraint in Eq.~\ref{eq:triplet}, a common approach is to minimize the following hinge loss:
\begin{aligns} \label{eq:triplet_hinge}
\Big[ \| \f_i^p - \f^+ \|^2 - \| \f_i^p - \f^- \|^2 + m \Big]_+.
\end{aligns}
Despite being broadly used, optimizing Eq.~\ref{eq:triplet_hinge} using standard triplet sampling leads to slow convergence and unstable performance in practice.
Inspired by the recent advance in metric learning, we enforce each of the three constraints by minimizing the N-pair loss\footnote{It is worth to point out that the implementation of MAMC is independent to the use of N-pair loss, as MAMC is a general framework that can be combined with other triplet-based metric learning loss as well.
The N-pair loss is taken as a reference because of its robustness and good convergence in practice.}~\cite{sohn2016improved},
\begin{aligns}
  L^{np} = \frac{1}{N} \sum_{\f_i^p \in \cB} \Big\{ \sum_{\f^+ \in \cP}\log\Big(1 + \sum_{\f^- \in \cN}\exp(\f_{i}^{pT} \f^- - \f_{i}^{pT} \f^+) \Big) \Big\}.
\end{aligns}

In general, for each training batch $\cB$, MAMC jointly minimizes the softmax loss and the N-pair loss with a weight parameter $\lambda$:
\begin{aligns}
  L^{mamc} = L^{softmax} + \lambda \Big( L^{np}_{sasc} + L^{np}_{sadc} + L^{np}_{dasc} \Big).
\end{aligns}

Given a batch of $N$ images and $P$ parts, MAMC is able to generate $2(PN-1)+4(N-1)^2(P-1)+4(N-1)(P-1)^2$ constraints of three types (Eq. \ref{eq:sasc} to Eq. \ref{eq:dasc}), while the N-pair loss can only produce $N-1$.
To put it in perspective, we are able to generate $130\times$ more constraints than N-pair loss with the same data under the normal setting where $P = 2$ and $N = 32$.
This implies that MAMC leverages much richer correlations among the samples, and is able to obtain better convergence than either triplet or N-pair loss.


\section{The Dogs-in-the-Wild Dataset}

Large image datasets (such as ImageNet~\cite{deng2009imagenet}) with high-quality annotations enables the dramatic development in visual recognition.
However, most datasets for fine-grained recognition are out-dated, non-natural and relatively small (as shown in Table~\ref{table:exp_datasets}).
Recently, there are several attempts such as Goldfinch~\cite{krause2016unreasonable} and the iNaturalist Challenge~\cite{van2017inaturalist} in building large-scale fine-grained benchmarks.
However, there still lacks a comprehensive dataset with large enough data volume, highly accurate data annotation, and full tag coverage of common dog species.
We hence introduce the Dogs-in-the-Wild dataset with 299,458 images of 362 dog categories, which is 15$\times$ larger than Stanford Dogs~\cite{khosla2011novel}.
We generate the list of dog species by combining multiple sources (\eg, Wikipedia), and then crawl the images with search engines (\eg, Google, Baidu).
The label of each image is then checked with crowd sourcing.
We further prune small classes with less than 100 images, and merge extremely similar classes by applying confusion matrix and manual validation.
The whole annotation process is conducted three times to guarantee the annotation quality.
Last but not least, since most of the experimental baselines are pre-trained on ImageNet, which has substantial category overlap with our dataset, we exclude any image of ImageNet from our dataset for fair evaluation.
This dataset will be released upon acceptance.

\begin{figure}[t]
  \centering
  \includegraphics[width=0.9\textwidth]{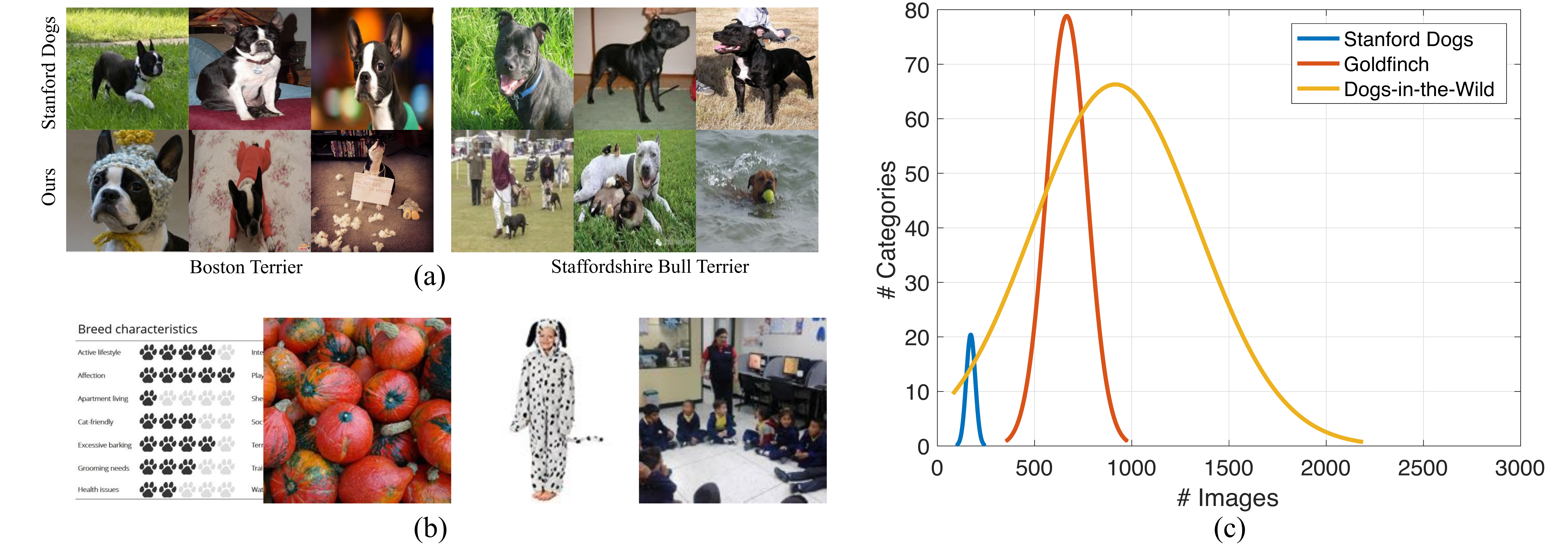}
  \caption{Qualitative and quantitative comparison of dog datasets. (a) Example images from Stanford Dogs and Dogs-in-the-Wild; (b) Common bad cases from Goldfinch that are completely non-dog. (c) Images per category distribution.}
  \label{fig:compare_dogs}
  \centering
\end{figure}

Figure~\ref{fig:compare_dogs}a and Figure~\ref{fig:compare_dogs}b qualitatively compare our dataset with the two most relevant benchmarks, Stanford Dogs~\cite{khosla2011novel} and the dog section of Goldfinch~\cite{krause2016unreasonable}.
It can be seen that our dataset is more challenging in two aspects:
(1) The intra-class variation of each category is larger.
For instance, almost all common patterns and hair colors of Staffordshire Bull Terriers are covered in our dataset, as illustrated in Figure~\ref{fig:compare_dogs}a.
(2) More surrounding environment types are covered, which includes but is not limited to, natural scenes, indoor scenes and even artificial scenes; and the dog itself could either be in its natural appearance or dressed up, such as the first Boston Terrier in Figure~\ref{fig:compare_dogs}a.
Another feature of our dataset is that all of our images are manually examined to minimize annotation errors.
Although Goldfinch has comparable class number and data volume, it is common to find noisy images inside, as shown in Figure~\ref{fig:compare_dogs}b.

We then demonstrate the statistics of the three datasets in Figure~\ref{fig:compare_dogs}c and Table~\ref{table:exp_datasets}.
It is observed that our dataset is significantly more imbalanced in term of images per category, which is more consistent with real-life situations, and notably increases the classification difficulty. Note that the curves in Figure~\ref{fig:compare_dogs}c are smoothed for better visualization.
On the other hand, the average images per category of our dataset is higher than the other two datasets, which contributes to its high intra-class variation, and makes it less vulnerable to overfitting.

\begin{table}[t]
   \small
    \begin{center}
    \makebox[\linewidth]{\resizebox{0.8\linewidth}{!}{
      \begin{tabular}{lcccc}
        \toprule
	Dataset & \#Class & \#Train & \#Test & \#Avg. Train/Class\\
        \midrule
	CUB-200-2011 & 200 & 5,994 & 5,794 & 30\\
	Stanford Dogs & 120 & 12,000 & 8,580 & 100\\
	Stanford Cars & 196 & 8,144 & 8,041 & 42\\
	Goldfinch & 515 & 342,632 & - & 665\\
	\bf{Dogs-in-the-Wild} & \bf{362} & \bf{258,474} & \bf{40,984} & \bf{714}\\
        \bottomrule
      \end{tabular}}}
  \end{center}
  \vspace{-.1in}
  \caption{Statistics of the related datasets.}
  \label{table:exp_datasets}
\end{table}

\section{Experimental Results}
\label{experimental_results}

We conduct our experiments on four fine-grained image recognition datasets, including three publicly available datasets CUB-200-2011~\cite{welinder2010caltech}, Stanford Dogs~\cite{khosla2011novel} and Stanford Cars~ \cite{krause20133d}, and the proposed Dogs-in-the-Wild dataset.
The detailed statistics including class numbers and train/test distributions are summarized in Table~\ref{table:exp_datasets}.
We adopt top-1 accuracy as the evaluation metric.

In our experiments, the input images are resized to 448$\times$448 for both training and testing.
We train on each dataset for 60 epochs; the batch size is set to 10 (N=5), and the base learning rate is set to 0.001, which decays by 0.96 for every 0.6 epoch.
The reduction ratio $r$ of $\W_1^p$ and $\W_2^p$ in Eq. \ref{eq:w1w2} is set to 16 in reference to~\cite{hu2017squeeze}.
The weight parameter $\lambda$ is empirically set to 0.5 as it achieves consistently good performances.
And for the FC layers, we set the channels $C=2048$ and $D=1024$.
Our method is implemented with Caffe \cite{jia2014caffe} and one Tesla P40 GPU.


\vspace{-.1in}
\subsection{Ablation Analysis}
To fully investigate our method, Table~\ref{table:results_abl} provides a detailed ablation analysis on different configurations of the key components.

\textbf{Base networks.} To extract convolutional feature before the OSME module, we choose VGG-19 \cite{simonyan2014very}, ResNet-50 and ResNet-101 \cite{he2016deep} as our candidate baselines. Based on Table~\ref{table:results_abl}, ResNet-50 and ResNet-101 are selected given their good balance between performance and efficiency.
We also note that although a better ResNet-50 baseline on CUB is reported in \cite{li2017dynamic} (84.5\%), it is implemented in Torch~\cite{collobert2011torch7} and tuned with more advanced data augmentation (\eg, color jittering, scaling).
Our baselines, on the other hand, are trained with simple augmentation (\eg, mirror and random cropping) and meet the Caffe baselines of other works, such as 82.0\% in \cite{liu2016fully} and 78.4\% in \cite{cui2017kernel}.

\textbf{Importance of OSME.} OSME is important in attending discriminative regions.
For ResNet-50 without MAMC, using OSME solely with $P=2$ can offer 3.2\% performance improvement compared to the baseline (84.9\% vs. 81.7\%).
With MAMC, using OSME boosts the accuracy by 0.5\% than without OSME (using two independent FC layers instead, 86.2\% vs. 85.7\%).
We also notice that two attention regions ($P=2$) lead to promising results, while more attention regions ($P=3$) provide slightly better performance.


\textbf{MAMC constraints.} Applying the first MAMC constraint (Eq.~\ref{eq:sasc}) achieves 0.5\% better performance than the baseline with ResNet-50 and OSME.
Using all of the three MAMC constraints (Eq.~\ref{eq:sasc} to Eq.~\ref{eq:dasc}) leads to another 0.8\% improvement.
This indicates the effectiveness of each of the three MAMC constraints.

\textbf{Complexity.} Compared with the ResNet-50 baseline, our method provides significantly better result (+4.5\%) with only 30\% more time, while a similar method \cite{fu2017look} offers less optimal result but takes $3.6\times$ more time than ours.

\begin{table}[p]
  \small
  \begin{minipage}[t]{\linewidth}
  \centering
  \begin{subtable}[t]{\textwidth}
    \begin{tabular*}{\textwidth}{lcccc}
      \toprule
      Method & $\#$Attention($P$) & 1-Stage & Acc. & Time(ms)\\
      \midrule
      VGG-19 & - & \checkmark & 79.0 & 79.8 \\
      ResNet-50 & - & \checkmark & 81.7 & 48.6 \\
      ResNet-101 & - & \checkmark & 82.5 & 82.7 \\
      ResNet-50 + OSME & 2  & \checkmark & 84.9 & 63.3 \\
      RACNN \cite{fu2017look} & 3  & $\times$ & 85.3 & 229 \\
      ResNet-50 + OSME + MAMC (Eq.~\ref{eq:sasc}) & 2 & \checkmark & 85.4 & 63.3 \\
      ResNet-50 + FC + MAMC (Eq.~\ref{eq:sasc}$\sim$\ref{eq:dasc}) & 2  & \checkmark & 85.7 & 60.3 \\
      ResNet-50 + OSME + MAMC (Eq.~\ref{eq:sasc}$\sim$\ref{eq:dasc}) & 2  & \checkmark & 86.2 & 63.3 \\
      ResNet-50 + OSME + MAMC (Eq.~\ref{eq:sasc}$\sim$\ref{eq:dasc}) & 3 & \checkmark & 86.3  & 68.1 \\
      ResNet-101 + OSME + MAMC (Eq.~\ref{eq:sasc}$\sim$\ref{eq:dasc}) & 2 & \checkmark & \bf86.5  & 102.1 \\
      \bottomrule
    \end{tabular*}
    \caption{Ablation analysis of our method on CUB-200-2011.}
    \label{table:results_abl}
  \end{subtable}
\end{minipage}
  \small
  \begin{minipage}[p]{0.48\linewidth}
    \centering
    \begin{subtable}[t]{\textwidth}
    \begin{tabular}[t]{lccc}
      \toprule
      Method & Anno. & 1-Stage & Acc.\\
      \midrule
	DVAN \cite{zhao2016diversified} & $\times$ & $\times$ & 79.0\\
	DeepLAC \cite{lin2015deep} & \checkmark & \checkmark & 80.3\\
	NAC \cite{simon2015neural} & $\times$ & \checkmark & 81.0\\
	Part-RCNN \cite{zhang2014part} & \checkmark & $\times$ & 81.6\\
	MG-CNN \cite{wang2015multiple} & $\times$ & $\times$ & 81.7\\
	ResNet-50 \cite{he2016deep} & $\times$ & \checkmark & 81.7\\
	PA-CNN \cite{krause2015fine} & \checkmark & \checkmark & 82.8\\
	RAN \cite{wang2017residual} & $\times$ & $\times$ & 82.8\\
	MG-CNN \cite{wang2015multiple} & \checkmark & $\times$ & 83.0\\
	B-CNN \cite{lin2015bilinear} & $\times$ & $\times$ & 84.1\\
	ST-CNN \cite{jaderberg2015spatial} & $\times$ & $\times$ & 84.1\\
	FCAN \cite{liu2016fully} & $\times$ & \checkmark & 84.3\\
	PDFR \cite{zhang2016picking} & $\times$ & $\times$ & 84.5\\
	ResNet-101 \cite{he2016deep} & $\times$ & \checkmark & 84.5\\
	FCAN \cite{liu2016fully} & \checkmark & \checkmark & 84.7\\
	SPDA-CNN \cite{zhang2016spda} & \checkmark & \checkmark & 85.1\\
	RACNN \cite{fu2017look} & $\times$ & $\times$ & 85.3\\
	PN-CNN \cite{branson2014bird} & \checkmark & $\times$ & 85.4\\
	RAM \cite{li2017dynamic} & $\times$ & $\times$ & 86.0\\
	MACNN \cite{zheng2017learning} & $\times$ & \checkmark & \bf86.5\\
	\midrule
        	Ours (ResNet-50) & $\times$ & \checkmark & 86.2\\
	Ours (ResNet-101) & $\times$ & \checkmark & \bf86.5\\
        \bottomrule
      \end{tabular}
      \caption{CUB-200-2011.} \label{table:results_cub}
    \end{subtable}
      \begin{subtable}[t]{\textwidth}
  	\begin{tabular}{lccc}
	\toprule
          Method & Anno. & 1-Stage & Acc.\\
	\midrule
	PDFR \cite{zhang2016picking} & $\times$ & $\times$ & 72.0\\
	ResNet-50 \cite{he2016deep} & $\times$ & \checkmark & 81.1\\
	DVAN \cite{zhao2016diversified} & $\times$ & $\times$ & 81.5\\
	RAN \cite{wang2017residual} & $\times$ & $\times$ & 83.1\\
	FCAN \cite{liu2016fully} & $\times$ & \checkmark & 84.2\\
	ResNet-101 \cite{he2016deep} & $\times$ & \checkmark & 84.9\\
	RACNN \cite{fu2017look} & $\times$ & $\times$ & \bf87.3\\
	\midrule
	Ours (ResNet-50) & $\times$ & \checkmark & 84.8\\
	Ours (ResNet-101) & $\times$ & \checkmark & 85.2\\
	\bottomrule
	\end{tabular}
        \caption{Stanford Dogs.}
        \label{table:results_stanford_dogs}
      \end{subtable}
 \end{minipage}
 \begin{minipage}[p]{0.48\linewidth}
      \centering
      \begin{subtable}[t]{\textwidth}
	\begin{tabular}{lccc}
	\toprule
         Method & Anno. & 1-Stage & Acc.\\
	\midrule
	DVAN \cite{zhao2016diversified} & $\times$ & $\times$ & 87.1\\
	FCAN \cite{liu2016fully} & $\times$ & \checkmark & 89.1\\
	ResNet-50 \cite{he2016deep} & $\times$ & \checkmark & 89.8\\
	RAN \cite{wang2017residual} & $\times$ & $\times$ & 91.0\\
	B-CNN \cite{lin2015bilinear} & $\times$ & $\times$ & 91.3\\
	FCAN \cite{liu2016fully} & \checkmark & \checkmark & 91.3\\
	ResNet-101 \cite{he2016deep} & $\times$ & \checkmark & 91.9\\
	RACNN \cite{fu2017look} & $\times$ & $\times$ & 92.5\\
	PA-CNN \cite{krause2015fine} & \checkmark & \checkmark & 92.8\\
	MACNN \cite{zheng2017learning} & $\times$ & \checkmark & 92.8\\
	\midrule
	Ours (ResNet-50) & $\times$ & \checkmark & 92.8\\
	Ours (ResNet-101) & $\times$ & \checkmark & \bf93.0\\
	\bottomrule
	\end{tabular}
        \caption{Stanford Cars.}
        \label{table:results_stanford_cars}
      \end{subtable}
      \begin{subtable}[b]{\textwidth}
	\begin{tabular}{lccc}
	\toprule
	Method & Anno. & 1-Stage & Acc.\\
	\midrule
	ResNet-50 \cite{he2016deep} & $\times$ & \checkmark & 74.4\\
	ResNet-101 \cite{he2016deep} & $\times$ & \checkmark & 75.6\\
	RAN \cite{wang2017residual} & $\times$ & $\times$ & 75.7\\
	RACNN \cite{fu2017look} & $\times$ & $\times$ & 76.5\\
	\midrule
	Ours (ResNet-50) & $\times$ & \checkmark & 77.9\\
	Ours (ResNet-101) & $\times$ & \checkmark & \bf78.5\\
	\bottomrule
	\end{tabular}
        \caption{Dogs-in-the-Wild.}
        \label{table:results_dogs-in-the-wild}
      \end{subtable}
      \end{minipage}
      \caption{Experimental results. ``Anno.'' stands for using extra annotation (bounding box or part) in training. ``1-Stage'' indicates whether the training can be done in one stage. ``Acc.'' denotes the top-1 accuracy in percentage.}
\end{table}

\vspace{-.1in}
\subsection{Comparison with State-of-the-Art}

In reference to \cite{fu2017look}, we select $18$ baselines as shown in Table~\ref{table:results_cub}.
Quantitative experimental results on the four datasets are shown in Table~\ref{table:results_cub}-\ref{table:results_dogs-in-the-wild}.

We first analyze the results on the CUB-200-2011 dataset in Table~\ref{table:results_cub}.
It is observed that with ResNet-101, our method achieves the best overall performance (tied with MACNN) against state-of-the-art.
Even with ResNet-50, our method exceeds the second best method using extra annotation (PN-CNN) by 0.8\%, and exceeds the second best method without extra annotation (RAM) by 0.2\%.
The fact that our method outperforms all of the methods with extra annotation demonstrates that good results are not necessarily linked with high costs.
For the weakly supervised methods without extra annotation, PDFR and MG-CNN conduct feature combination from multiple scales, and RACNN is trained with multiple alternative stages, while our method is trained with only one stage to obtain all the required features.
Yet our method outperforms all of the the three methods by 2.0\%, 4.8\% and 1.2\%, respectively.
The methods B-CNN and RAN share similar multi-branch ideas with the OSME in our method, where B-CNN connects two CNN features with outer product, and RAN combines the trunk CNN feature with an additional attention mask.
Our method, on the other hand, applies the OSME for multi-attention feature extraction in one step, which surpasses B-CNN and RAN by 2.4\% and 3.7\%, respectively.

Our method exhibits similar performances on the Stanford Dogs and Stanford Cars datasets, as shown in Table~\ref{table:results_stanford_dogs} and Table~\ref{table:results_stanford_cars}.
On Stanford Dogs, our method exceeds all of the comparison methods except RACNN, which requires multiple stages for feature extraction and is hard to be trained end-to-end.
On Stanford Cars, our method obtains 93.0\% accuracy, outperforming all of the comparison methods. It is worth noting that compared with the methods exploiting multi-scale or multi-stage information like DVAN and RAN, our method achieves significant improvements with only one feedforward stage for multi-attention multi-class feature extraction, which further validates the effectiveness and efficiency of our method.

Finally, on the Dogs-in-the-Wild dataset, our method still achieves the best result with remarkable margins. Since this dataset is newly proposed, the results in Table~\ref{table:results_dogs-in-the-wild} can be used as baselines for future explorations. Moreover, by comparing the overall performances in Table~\ref{table:results_stanford_dogs} and Table~\ref{table:results_dogs-in-the-wild}, we find that the accuracies on Dogs-in-the-wild are significantly lower than those on Stanford Dogs, which witness the relatively higher classification difficulty of this dataset.

\begin{figure*}[t]
	\centering
	\includegraphics[width=\textwidth]{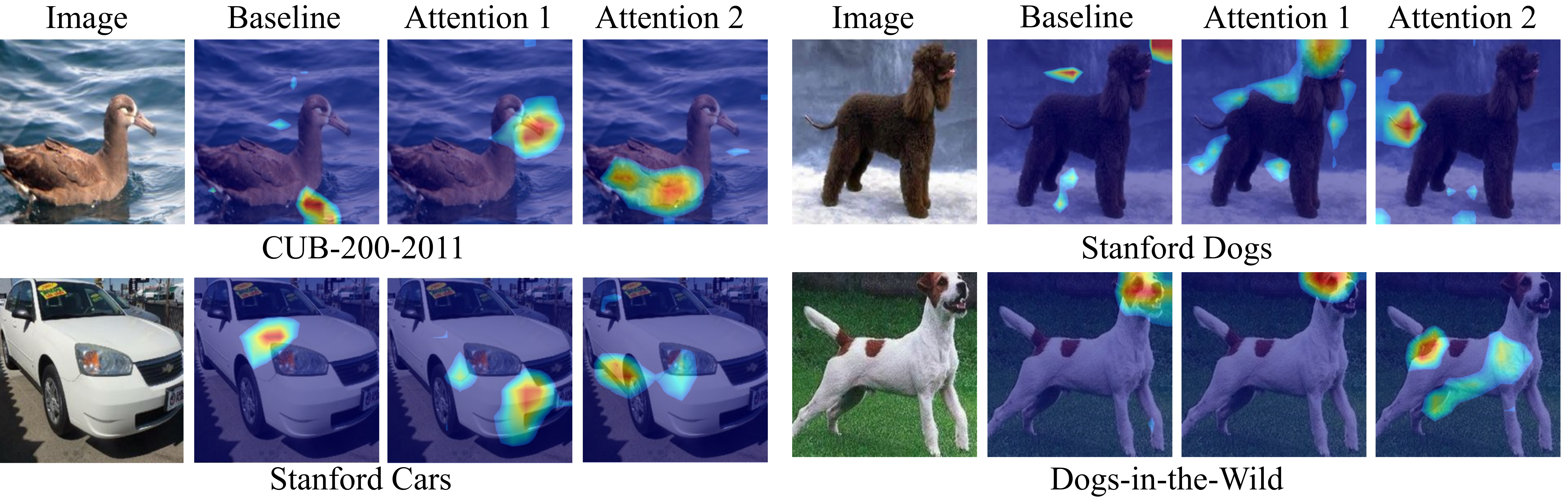}
	\caption{Visualization of the attention regions detected by the OSME. For each dataset, the first column shows the input image, the second column shows the heatmap from the last conv layer of the baseline ResNet-101; the third and fourth columns show the heatmaps of the two detected attention regions via OSME.}
	\label{fig:visualization}
        \vspace{-.1in}
\end{figure*}

By adopting our network with ResNet-101, we visualize the $\bS^p$ in Eq. \ref{eq:Sp} of each OSME branch (which corresponds to an attention region) as its channel-wise average heatmap, as shown in the third and fourth columns of Figure~\ref{fig:visualization}, .
In comparison, we also show the outputs of the last conv layer of the baseline network (ResNet-101) as heatmaps in the second column.
It is seen that the highlighted regions of OSME outputs reveal more meaningful parts than those of the baseline, that we humans also rely on to recognize the fine-grained label, \eg, the head and wing for birds, the head and tail for dogs, and the headlight/grill and frame for cars.

\vspace{-.1in}
\section{Conclusion}
\label{conclustion}

In this paper, we propose a novel CNN with the multi-attention multi-class constraint (MAMC) for fine-grained image recognition.
Our network extracts attention-aware features through the one-squeeze multi-excitation (OSME) module, supervised by the MAMC loss that pulls positive features closer to the anchor, while pushing negative features away.
Our method does not require bounding box or part annotation, and can be trained end-to-end in one stage.
Extensive experiments against state-of-the-art methods exhibit the superior performances of our method on various fine-grained recognition tasks on birds, dogs and cars.
In addition, we have collected and will release the Dogs-in-the-Wild, a comprehensive dog species dataset with the largest data volume, full category coverage, and accurate annotation compared with existing similar datasets.


\bibliographystyle{splncs}
\bibliography{references}
\end{document}